\documentclass[sigconf, nonacm, review]{acmart}
\usepackage{bm}
\usepackage{multirow}
\usepackage{booktabs}   
\usepackage{graphicx} 
\AtBeginDocument{%
  }

\begin{document}
\pagestyle{plain}
\settopmatter{printacmref=false} %
\title{TeMuDance: Contrastive Alignment-Based \\ Textual Control for 
Music-Driven Dance Generation}


\author{Xinran Liu$^{1}$, Diptesh Kanojia$^{1}$, Wenwu Wang$^{1}$, Zhenhua Feng$^{2}$}

\affiliation{
  \institution{$^{1}$University of Surrey, Guildford, Surrey, UK \quad $^{2}$Jiangnan University, Wuxi, Jiangsu, China}
  \country{}
}

\thanks{Preprint. Under review.}

\email{{xl01315, d.kanojia, w.wang}@surrey.ac.uk, fengzhenhua@jiangnan.edu.cn}




\begin{abstract}
Existing music-driven dance generation approaches have achieved strong realism and effective audio-motion alignment. However, they generally lack semantic controllability, making it difficult to guide specific movements through natural language descriptions. This limitation primarily stems from the absence of large-scale datasets that jointly align music, text, and motion for supervised learning of text-conditioned control. To address this challenge, we propose TeMuDance, a framework that enables text-based control for music-conditioned dance generation without requiring any manually annotated music--text--motion triplet dataset. TeMuDance introduces a motion-centred bridging paradigm that leverages motion as a shared semantic anchor to align disjoint music–dance and text–motion datasets within a unified embedding space, enabling cross-modal retrieval of missing modalities for end-to-end training. A lightweight text control branch is then trained on top of a frozen music-to-dance diffusion backbone, preserving rhythmic fidelity while enabling fine-grained semantic guidance. To further suppress noise inherent in the retrieved supervision, we design a dual-stream fine-tuning strategy with confidence-based filtering. We also propose a novel task-aligned metric that quantifies whether textual prompts induce the intended kinematic attributes under music conditioning. 
Extensive experiments demonstrate that TeMuDance achieves competitive dance quality while substantially improving text-conditioned control over existing methods. 
\end{abstract}

\maketitle

\vspace{0.5cm}
\section{Introduction}

\begin{figure}[!t]
  \centering
  \includegraphics[width=.5\textwidth]{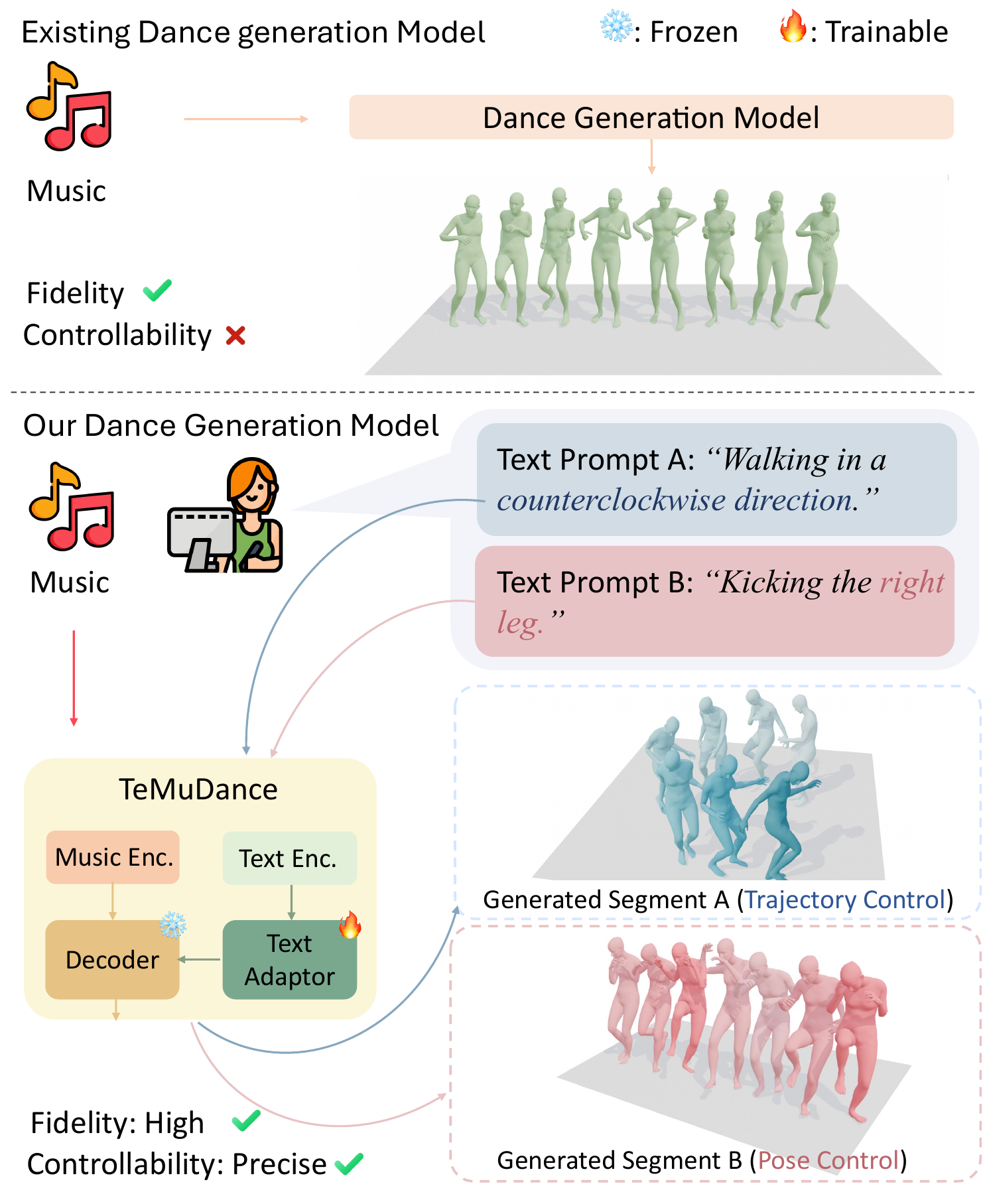}
  \caption{The proposed TeMuDance method is able to generate dances conditioned on music and text jointly, producing sequences that are both rhythmically aligned and semantically controllable.}
  \label{fig:top}
\end{figure}

In recent years, the media production industry has fueled a strong demand for automated, high-fidelity character animation~\cite{mourot2022survey,zhu2023human}. As a complex form of expressive motion, music-driven 3D dance generation has emerged as an important research domain, aiming to enable virtual characters to synthesize realistic movements from music~\cite{sun2020deepdance,li2021ai}.
Despite the impressive realism achieved by current dance generation approaches~\cite{kim2022brand,siyao2022bailando,tseng2023edge, li2024lodge}, a critical limitation remains: the lack of fine-grained semantic control. Most existing methods rely on coarse conditioning mechanisms and struggle to consistently follow explicit, intention-aligned instructions, which significantly reduces their practicality in real-world production settings.

To enable controllability, some methods employ coarse cues such as global genre labels~\cite{liu2025gcdance}, which provide only high-level stylistic guidance and cannot convey complex semantic intents. Finer-grained approaches, including text-guided editing~\cite{zhang2025danceeditor} and discrete codebook-based motion representations~\cite{gong2023tm2d}, introduce their own trade-offs: editing pipelines are constrained by the supervision distributions available in current datasets, limiting generalisation to unseen instructions such as spatial trajectories; discrete quantization can disconnect semantic actions from musical rhythm, preventing user-specified movements from being naturally integrated with the accompaniment.
In essence, the challenge arises from the disjoint nature of existing datasets. Music–dance datasets provide rhythmic alignment but lack textual annotations, while text–motion datasets provide language supervision but without accompanying music. The absence of music–text–motion triplets therefore prevents existing models from jointly learning rhythmic coherence and semantic control.

To bridge this gap, we introduce TeMuDance, which enables text-based control for music-conditioned 3D dance generation, as shown in Figure~\ref{fig:top}. Specifically, our model learns text controllability without requiring any paired music–text–motion supervision. At its core, TeMuDance introduces a motion-centred bridging mechanism that leverages motion as a shared semantic anchor to align separate music–dance and text–motion datasets within a unified embedding space. This unified representation enables cross-modal retrieval of missing modalities, providing end-to-end supervision for training a jointly conditioned generator while avoiding the artifacts associated with discrete quantization.

To preserve high-quality music-driven dance generation while incorporating textual control, we first pretrain a music-to-dance generation model and freeze it as the backbone, retaining its strong rhythmic alignment and physical realism. We then attach a text-conditioned control branch that injects textual features into intermediate layers, steering generation toward the desired semantics without modifying the backbone parameters. This design enables TeMuDance to achieve fine-grained textual control while maintaining music-synchronized dance quality.

In addition, we employ a dual-stream training strategy that combines mutual dataset augmentation with confidence-based noise filtering, thereby suppressing noise from pseudo annotations and enhancing the precision of semantic control. Moreover, existing text–motion evaluation metrics are designed for text-only generation and cannot faithfully assess text controllability under music conditioning. We therefore introduce a dedicated protocol grounded in kinematic predicates that directly verifies the successful execution of text-specified actions in the generated dance.

Overall, the main contributions of TeMuDance are summarised as follows.

(1) We propose TeMuDance, a novel framework that enables textual control in music-driven dance generation without requiring any music–text–dance triplet dataset.

(2) We introduce motion-centred bridging that aligns disjoint music–dance and text–motion pairs into a shared latent space, enabling semantic concepts learned from text–motion data to transfer to music-conditioned dance synthesis.

(3) We design a dual-stream training strategy with confidence-based filtering to suppress noisy pseudo annotations arising from cross-dataset inference while allowing the two data streams to reinforce each other.

(4) We propose Kinematic Primitive Success (KPS), a task-aligned metric that measures whether textual conditioning successfully drives the expected kinematic patterns in generated dances.

\section{Related Work}

\subsection{3D Human Motion Synthesis}
Traditionally, 3D human skeletal motion prediction relies on physics-based methods that explicitly model kinematics, dynamics, and physical constraints of the human body, which are often computationally complex and unstable~\cite{loi2023machine}. 
More recently, learning-based approaches leverage large-scale datasets to enable more efficient and accurate prediction of 3D motion trajectories. 
Specifically, early efforts primarily employ RNNs for this task~\cite{martinez2017human, li2018convolutional, liu2019towards}. 
However, RNN-based models are susceptible to error accumulation, which can lead to discontinuities in predicted motion sequences~\cite{gui2018adversarial}. 
Ma et al.~\cite{ma2022progressively} propose a network composed of spatial dense GCNs and temporal dense GCNs, which alternates between spatial and temporal modules to extract spatiotemporal features over the global receptive field. 
Aksan et al.~\cite{aksan2021spatio} utilise a self-attention mechanism to learn high-dimensional joint embeddings and generate temporally coherent poses.

The Motion Diffusion Model (MDM)~\cite{tevet2022humanmotiondiffusionmodel} is the first to apply classifier-free diffusion to human motion generation, which inspires many subsequent diffusion-based approaches.
MotionFix~\cite{athanasiou2024motionfix} conditions diffusion models on both source motion and edit text for seamless motion edits. 
Although prior work improves motion quality and diversity, dance generation remains challenging because it requires both precise beat synchrony and consistent genre-aligned style.

\subsection{Music Driven Dance Generation}
Early studies~\cite{shiratori2006dancing, ofli2008audio, fukayama2015music} consider this task as a similarity-based retrieval problem. With the advent of deep learning, it is reframed as a supervised motion prediction problem, leveraging architectures such as CNN~\cite{holden2016deep,holden2015learning}, RNN~\cite{butepage2017deep,chiu2019action,du2019bio}, and Transformer~\cite{fan2022bi,huang2022genre,li2022danceformer}. 
However, these frame-by-frame prediction approaches often face challenges such as error accumulation and motion freezing~\cite{zhuang2022music2dance}. 

Recent research shifts to a generative pipeline. While methods based on VQ-VAE~\cite{gong2023tm2d,siyao2022bailando} have achieved outstanding performance, these systems are highly complex and involve multiple sub-networks. 
EDGE~\cite{tseng2023edge} is the first method that employs a diffusion-based framework, featuring a single-model design optimised for a single objective. It also introduces a novel evaluation approach focusing on physical plausibility. Despite this progress in generation quality, a critical limitation persists: the lack of fine-grained semantic control.

\begin{figure*}[!t]
  \centering
  \includegraphics[width=1\textwidth]{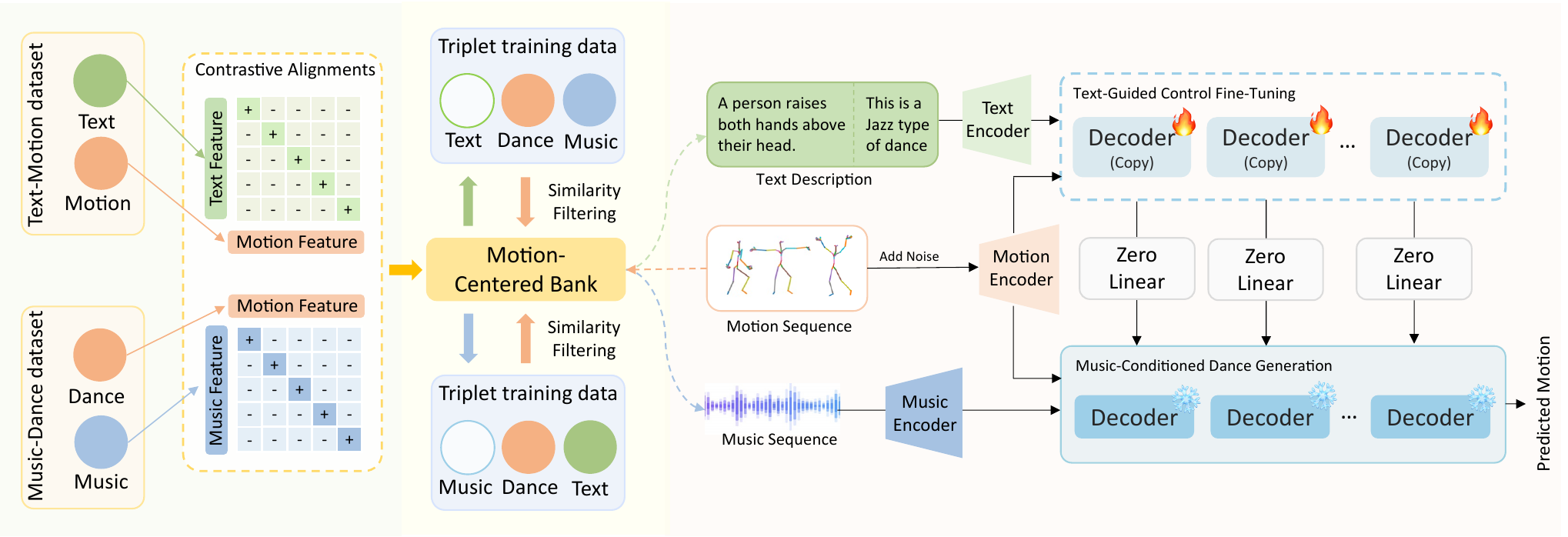}
  \caption{An overview of TeMuDance. We learn a motion-centred bank by contrastively aligning disjoint text–motion and music–dance datasets in a shared motion space, enabling similarity-filtered modality completion to form pseudo triplets. For generation, a pretrained diffusion Transformer is frozen as the backbone, while a text control branch steers denoising and produces rhythm-aligned, semantically controllable dances. }
  \label{f2}
\end{figure*}

\subsection{Controllable Dance Generation}
To enable controllability in dance generation, several approaches~\cite{huang2022genre, liu2025gcdance} utilise discrete genre embeddings to achieve coarse-grained style control. While effective for global stylization, these label-driven methods lack the granularity to specify concrete motion details.

To enable flexible semantic control, recent research increasingly explores text-driven generation and editing. For example, DanceEditor~\cite{zhang2025danceeditor} proposes an iterative editing paradigm that leverages language guidance to progressively revise motions, enabling targeted modifications beyond coarse style switching. TM2D~\cite{gong2023tm2d} takes a step towards finer control by introducing action-annotated data and explicitly modeling controllable action units. However, its VQ-VAE discretisation can hinder smooth transitions and seamless choreographic integration. In parallel, general-purpose multimodal motion generators, such as UniMuMo~\cite{yang2025unimumo}, MotionAnything~\cite{zhang2025motion}, and DanceChat~\cite{wang2025dancechat}, aim to unify motion synthesis under diverse conditioning signals, including text and music, within a single backbone. Despite richer conditioning, these generalist frameworks often treat text as a global cue, leading the model to follow instructions at the pose or clip level rather than to coherent choreography-level control.

\begin{figure}[!t]
  \centering
  \includegraphics[width=0.5\textwidth]{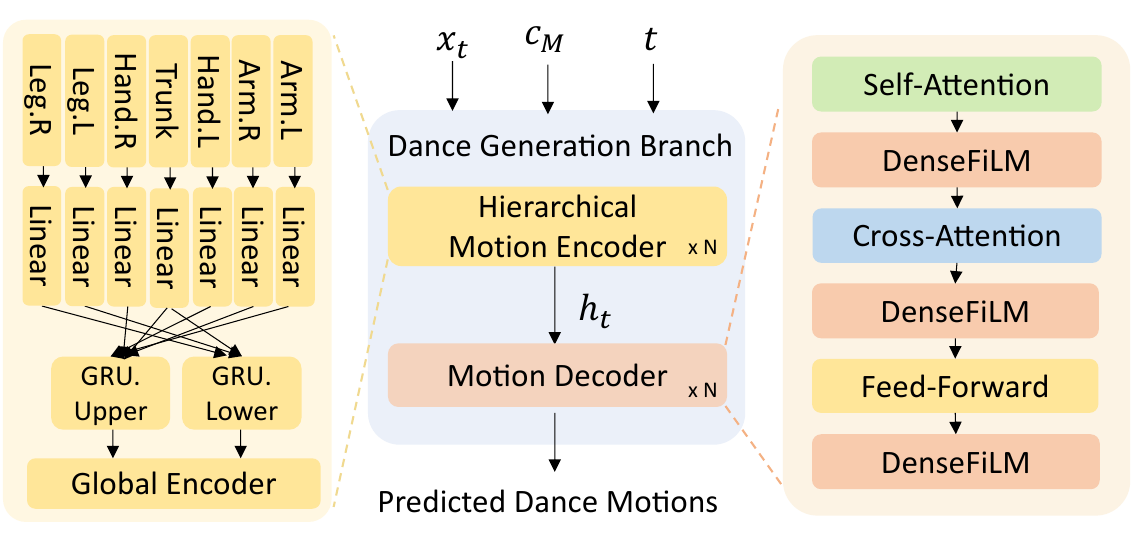}
  \caption{Architecture of the music-conditioned diffusion dance generator. 
}
  \label{fig:backbone_arch}
\vspace{-0.3cm}
\end{figure}

\section{The Proposed TeMuDance Method}
We present the overall framework of TeMuDance in Figure~\ref{f2}, which comprises a high-fidelity music-driven dance generation backbone, a text-conditioned adapter that enables semantic control, and a motion-anchored bridging strategy for cross-modal alignment, thereby strengthening controllability while preserving motion quality.

\subsection{Music-conditioned Dance Generator}
\label{sec:backbone}
Our approach builds upon a pretrained diffusion-based music conditioned dance generator that maps a music segment to a temporally coherent $3$D dance sequence. Given a long music-dance pair, we partition it into $4$-second clips and uniformly sample $\mathit{k}$ segments per clip. Each segment is represented using the SMPL-X parameterization~\cite{loper2023smpl}. We denote a motion clip as $\bm{x} \in \mathbb{R}^{k \times F}$, where $F = 319$ denotes the dimensionality of the skeletal motion features. We provide details in Appendix~\ref{app:motion}.
The corresponding music clip is encoded into temporally aligned conditioning features $\bm{c}_M \in \mathbb{R}^{k \times C}$ using a pretrained music foundation model combined with low-level waveform descriptors following~\cite{liu2025gcdance}, where $C$ denotes the music feature dimension.

We employ a diffusion-based dance generation backbone following the Denoising Diffusion Probabilistic Model (DDPM)~\cite{ho2020ddpm}. At each training step, we sample a timestep $t$ and add noise to the clean motion clip $\bm{x}_0$ to obtain $\bm{x}_t$:
\begin{equation}
\bm{x}_t=\sqrt{\bar{\alpha}_t}\,\bm{x}_0+\sqrt{1-\bar{\alpha}_t}\,\bm{\epsilon}
\end{equation}
where $\bm{\epsilon}\sim\mathcal{N}(\bm{0},\bm{I})$ and $\bar{\alpha}_{t}\in(0,1)$ denotes a monotonically decreasing schedule. 

As shown in Figure~\ref{fig:backbone_arch}, the denoiser of the dance generation backbone comprises a Spatially Hierarchical Motion Encoder $E$ and a Denoising Decoder $D$.
To capture part-specific motion patterns while preserving whole-body coherence, the encoder $E$ partitions the input channels into $M=7$ body-part groups. Each group is processed by a hierarchical module to model local dynamics, followed by a fusion layer to capture inter-part dependencies, yielding the latent feature:
\begin{equation}
\bm{h}_t=E(\bm{x}_t)\in\mathbb{R}^{k\times H}
\end{equation}
where $H$ is the hidden feature dimension of the denoiser.
Subsequently, these features are fed into the Denoising Decoder $D$ to reconstruct the clean motion $\hat{\bm{x}}_0$. 
Each layer of $D$ comprises a self-attention mechanism for temporal modeling, a cross-attention mechanism that integrates the music features $\bm{c}_M$, and a feed-forward network modulated by the timestep $t$ through Feature-wise Linear Modulation (FiLM) layers~\cite{perez2018film}.
The network is trained to reconstruct the clean motion by the following objective:
\begin{equation}
\mathcal{L}_{\mathrm{diff}}=
\mathbb{E}_{\bm{x}_0,t}
\left[\left\|\bm{x}_0-D(\bm{h}_t,t,\bm{c}_M)\right\|_2^2\right]
\end{equation}

In addition to $\mathcal{L}_{\mathrm{diff}}$, following the settings of~\cite{tseng2023edge,tevet2022humanmotiondiffusionmodel}, we incorporate standard kinematic regularisers, including a joint position loss $\mathcal{L}_{\mathrm{joint}}$, a pose velocity and acceleration loss $\mathcal{L}_{\mathrm{vel}}$, and foot contact loss $\mathcal{L}_{\mathrm{contact}}$, to promote physically plausible and visually smooth motions:
\begin{equation}
\mathcal{L}_{\mathrm{dance}}
= \tau\!\left(
\mathcal{L}_{\mathrm{diff}},
\mathcal{L}_{\mathrm{joint}},
\mathcal{L}_{\mathrm{vel}},
\mathcal{L}_{\mathrm{contact}}
\right)
\end{equation}
where $\tau(\cdot)$ aggregates multiple loss terms into a scalar objective. During pretraining, we adopt Aligned Multi-Task Learning (Aligned-MTL)~\cite{senushkin2023independent}, which mitigates gradient conflicts among competing objectives, to stabilise the joint optimisation of these loss terms.

\subsection{Text-Guided Control Fine-Tuning}
\label{sec:control_module}

The pretrained backbone is conditioned solely on music. Our objective is to enable free-form textual steering while preserving the motion quality and rhythmic fidelity of the dance generator. To this end, inspired by ControlNet~\cite{zhang2023adding}, we construct a trainable text-conditioned control branch by duplicating the denoiser of the pretrained backbone. The original music-conditioned denoiser is kept frozen, while the control branch is trained to predict layer-wise residual signals that are injected into the corresponding blocks of the frozen denoiser. 

Given a text prompt, we use the BERT~\cite{devlin2019bert} text encoder to extract contextual features, which are projected to yield the text condition embedding $\bm{c}_E \in \mathbb{R}^{N \times H}$, where $N$ is the token sequence length.
The frozen denoiser includes $L$ stacked Transformer decoders $\{B^{(\ell)}\}_{\ell=0}^{L-1}$. We denote  $\bm{h}_t^{(\ell)}$ as the hidden state serving as the input to the $\ell$-th block, with $\bm{h}_t^{(0)} = E(\bm{x}_t)$. The control branch mirrors the structure of the first $K$ blocks of the backbone. Analogous to the backbone described in Sec.~\ref{sec:backbone}, the control blocks employ cross-attention layers to inject the condition embeddings. For the first $K$ blocks (i.e., $\ell=0,\dots,K-1$), the control branch predicts a residual $\bm{\Delta}^{(\ell)}$ that is injected into the corresponding frozen block:
\begin{align}
 \bm{\Delta}^{(\ell)} &= \mathcal{Z}^{(\ell)}\!\left(B'^{(\ell)}(\bm{h}_t^{(\ell)}, \bm{c}_E, t)\right) \\
    \bm{h}_t^{(\ell+1)} &= B^{(\ell)}(\bm{h}_t^{(\ell)}, \bm{c}_M, t) + \bm{\Delta}^{(\ell)}
\end{align}
where $B^{(\ell)}$ is the $\ell$-th block of the frozen backbone, and $B'^{(\ell)}$ is its trainable counterpart in the control branch. 
$\mathcal{Z}^{(\ell)}$ represents a zero-initialized linear projection layer. 
The zero-initialization ensures that $\bm{\Delta}^{(\ell)}$ starts at zero, making the generator function-preserving at the beginning of fine-tuning.
During fine-tuning, we update only the control branch and optimise it with the same dance objective as the backbone.
We denote this training loss as $\mathcal{L}_{\mathrm{text}}$.

\subsection{Motion-bridging Cross-Modal Alignment}
A key challenge in our setting is the absence of paired music--text--motion triplets. Since direct supervision is unavailable, we propose a \textit{motion-centred bridging framework} that operates in two stages. First, we use motion as a pivot to embed disjoint datasets into a shared latent space, establishing a unified foundation for cross-modal retrieval. Second, we introduce a dual-stream training strategy, balancing text controllability with the rhythmic fidelity of generated motion.

\subsubsection{Motion-Centred Latent Alignment}
\label{sec:alignment}
To enable cross-modal semantic transfer without paired music--text--motion triplets, we adopt a motion-centred contrastive formulation.
Specifically, we use the FineDance~\cite{li2023finedance} dataset for music--dance supervision and the HumanML3D~\cite{guo2022generating} dataset for text--motion supervision.
We then learn a unified embedding space with two contrastive streams and a motion-level regulariser to remain domain-consistent across datasets.

For the music--dance stream, we optimise a queue-based InfoNCE loss~\cite{he2020momentum}.
Given a paired sample $(\bm{c}_M, \bm{x}^{\mathrm{Da}}_0)$, we first apply temporal pooling operators $\rho_M(\cdot)$ and $\rho_X(\cdot)$ to aggregate the music and motion token sequences into global vectors, respectively.
These vectors are then mapped into a shared embedding space via learnable linear projectors $P_{\mathrm{mus}}$ and $P_{\mathrm{mot}}$.
The $\ell_2$-normalised query and key are defined as:
\begin{equation}
\begin{aligned}
\bm{q} &= \mathrm{norm}\!\left(P_{\mathrm{mus}}(\rho_M(\bm{c}_M))\right) \\
\bm{k} &= \mathrm{norm}\!\left(P_{\mathrm{mot}}(\rho_X(\bar{E}(\bm{x}^{\mathrm{Da}}_0)))\right)
\end{aligned}
\end{equation}
where $\bar{E}$ is an exponential moving average (EMA) copy of the motion encoder used to compute stable keys.
Let $\mathcal{Q}_{\mathrm{Da}}=[\bm{u}_1,\ldots,\bm{u}_{K_q}]\in\mathbb{R}^{D\times K_q}$ be a momentum-updated queue of $K_q$ negative motion keys, where $D$ is the projection dimension.
We minimise:
\begin{equation}
\begin{aligned}
\ell_{\mathrm{mus}}
=
-\,s_{\mathrm{mus}}\bm{q}^{\top}\bm{k}
+\log\!\Big(
\exp\!\big(s_{\mathrm{mus}}\bm{q}^{\top}\bm{k}\big)\\
+\sum_{j=1}^{K_q}\exp\!\big(s_{\mathrm{mus}}\bm{q}^{\top}\bm{u}_j\big)
\Big)\qquad\qquad
\end{aligned}
\end{equation}
where $s_{\mathrm{mus}}=\exp(\alpha_{\mathrm{mus}})$ and $\alpha_{\mathrm{mus}}$ is a learnable logit-scale parameter.
The final loss is $\mathcal{L}_{\mathrm{m2d}}=\mathbb{E}[\ell_{\mathrm{mus}}]$.

In parallel, for the text--motion stream, we use the same contrastive form $\mathcal{L}_{\mathrm{t2m}}$ to align text descriptions with their corresponding motion embeddings.

Although both streams share the motion encoder, the motion distributions of FineDance and HumanML3D are inherently different, training them independently can separate the two motion domains in the embedding space, breaking the semantic bridge between music and text.
To reduce domain drift, we regularise motion embeddings by aligning their batch-wise mean and covariance across the two domains:
\begin{equation}
\mathcal{L}_{\mathrm{bridge}} =
\left\lVert \bm{\mu}_{\mathrm{Da}} - \bm{\mu}_{\mathrm{Mo}} \right\rVert_2^2 +
\left\lVert \bm{\Sigma}_{\mathrm{Da}} - \bm{\Sigma}_{\mathrm{Mo}} \right\rVert_F^2
\end{equation}
where $\bm{\mu}$ and $\bm{\Sigma}$ denote the batch-wise mean vector and covariance matrix of motion embeddings, the subscripts $_{\mathrm{Da}}$ and $_{\mathrm{Mo}}$ denote the dance and motion domains of the FineDance and HumanML3D datasets, respectively, and $\lVert\cdot\rVert_F$ is the Frobenius norm. 

The overall alignment objective is
\begin{equation}
\mathcal{L}_{\mathrm{align}}=\mathcal{L}_{\mathrm{m2d}}+\mathcal{L}_{\mathrm{t2m}}+\lambda\,\mathcal{L}_{\mathrm{bridge}}
\end{equation}

\begin{figure}[!t]
  \centering
  \includegraphics[width=0.5\textwidth]{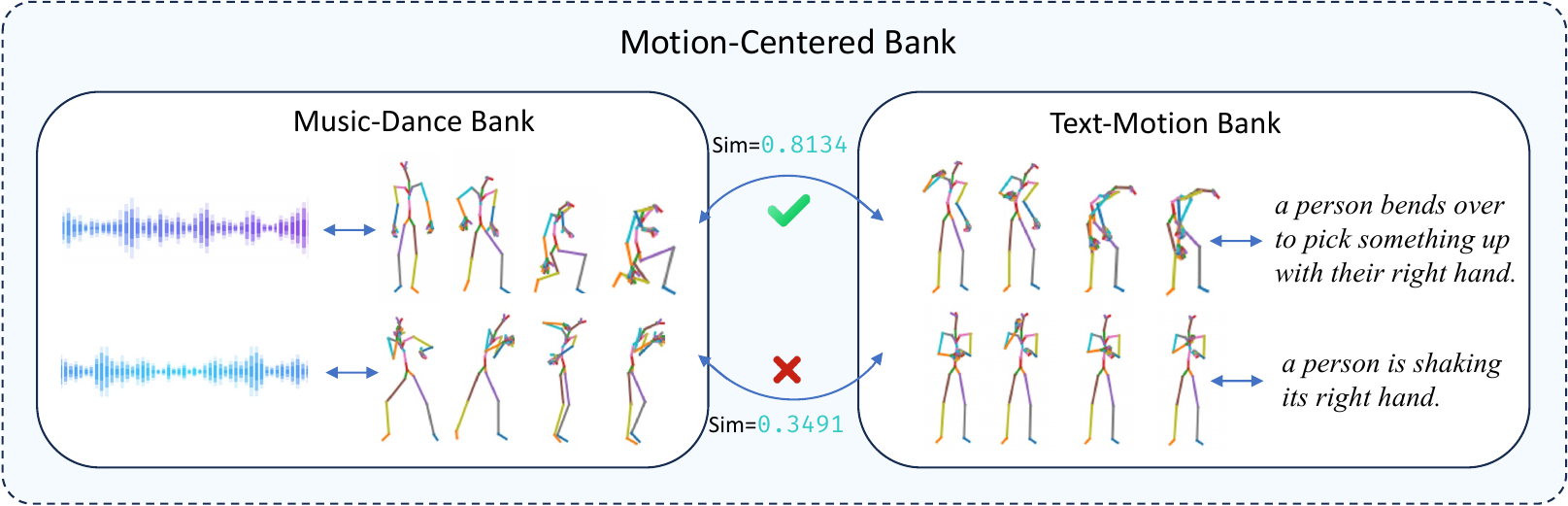}
  \caption{Illustration of the Motion-Centred Bank. 
}
  \vspace{-0.0cm}
  \label{fig:motion_bank}

\end{figure}

\subsubsection{Motion-Centred Dual-Stream Training}
\label{sec:dual_stream}

Although the contrastive alignment brings the unpaired datasets into a shared latent space, a key challenge in dual-stream fine-tuning remains: we aim to learn a jointly music--text conditioned generator, but the available supervision consists only of disjoint text-motion and music-dance pairs.
To address this missing-modality issue, we construct Motion-Centred Bank that enable motion-bridging cross-domain retrieval.
Specifically, we freeze the encoders and index the datasets into two Motion-Centred Bank, denoted as $\mathcal{B}_{\mathrm{MD}}$ (music-dance) and $\mathcal{B}_{\mathrm{TM}}$ (text-motion).

As illustrated in Figure~\ref{fig:motion_bank}, these banks serve as the foundation for cross-domain retrieval, enabling us to synthesize pseudo-triplets by imputing missing modalities. For a mini-batch sampled from the text-motion dataset, we have paired text and motion $(\bm{c}_E, \bm{x}^{\mathrm{Mo}}_0)$ but no music. We retrieve a rhythmically compatible music condition by querying the music-dance bank with the motion embedding. 
Concretely, we compute motion embeddings $ E(\bm{x}^{\mathrm{Mo}}_0)$, perform nearest-neighbour search in $\mathcal{B}_{\mathrm{MD}}$ with cosine similarity, and obtain the corresponding music features:

\begin{equation}
\tilde{\bm{c}}_M = {\mathcal{B}_{\mathrm{MD}}}(E(\bm{x}^{\mathrm{Mo}}_0))
\end{equation}
where matches falling below a similarity threshold are replaced by a null condition, ensuring that low-confidence pseudo annotations do not propagate into training. The acceptance rates and retrieval quality statistics are provided in Appendix~\ref{app:retrieval}.

This yields pseudo-triplets $(\bm{x}^{\mathrm{Mo}}_0, \tilde{\bm{c}}_M, \bm{c}_E)$ to train the text-control branch to steer denoising under music-compatible priors.
For a mini-batch sampled from the music--dance dataset, we analogously impute the missing text condition by querying the text--motion bank with the motion embedding to obtain $(\bm{x}^{\mathrm{Da}}_0, \bm{c}_M, \tilde{\bm{c}}_E)$.
To bridge the gap between specific motion semantics and global musical style, we construct a composite instruction by concatenating the retrieved description with coarse music genre tags, encouraging the control branch to follow both fine-grained actions and global style cues.

As illustrated in Figure~\ref{fig:dual_stream}, we fine-tune the joint denoiser $D_{\mathrm{joint}}$ by alternating between two streams utilizing the objectives defined in Sec.~\ref{sec:backbone} and Sec.~\ref{sec:control_module}. Specifically, the text--motion stream optimizes $\mathcal{L}_{\mathrm{text}}$ for semantic control, while the music--dance stream optimizes $\mathcal{L}_{\mathrm{dance}}$ to preserve the rhythmic prior. Accordingly, the final fine-tuning objective is formulated as a weighted combination:

\begin{figure}[!t]
  \centering
  \includegraphics[width=0.3\textwidth]{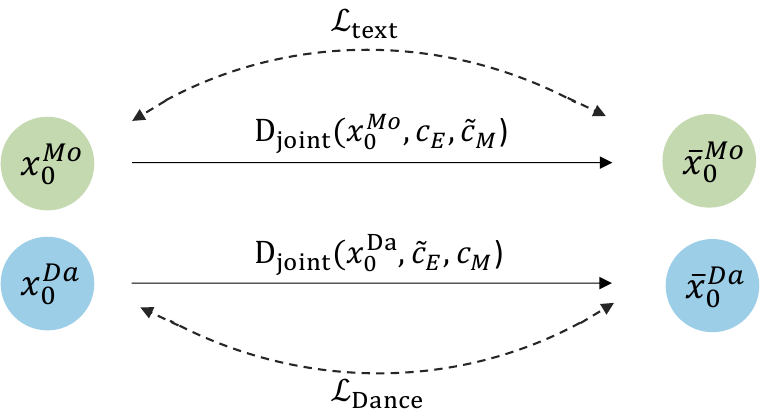}
  \caption{Visual pipeline of Dual-Stream Training. 
}
\vspace{-0.0cm}
  \label{fig:dual_stream}
\end{figure}

\begin{equation}
\begin{split}
\mathcal{L}_{\mathrm{ft}} = \, & (1-\lambda_p) \, \mathcal{L}_{\text{text}} + \lambda_p \, \mathcal{L}_{\text{dance}}
\end{split}
\end{equation}
where $\lambda_p$ is a trade-off hyperparameter.

\begin{figure}[!t]
  \centering
  \includegraphics[width=0.5\textwidth]{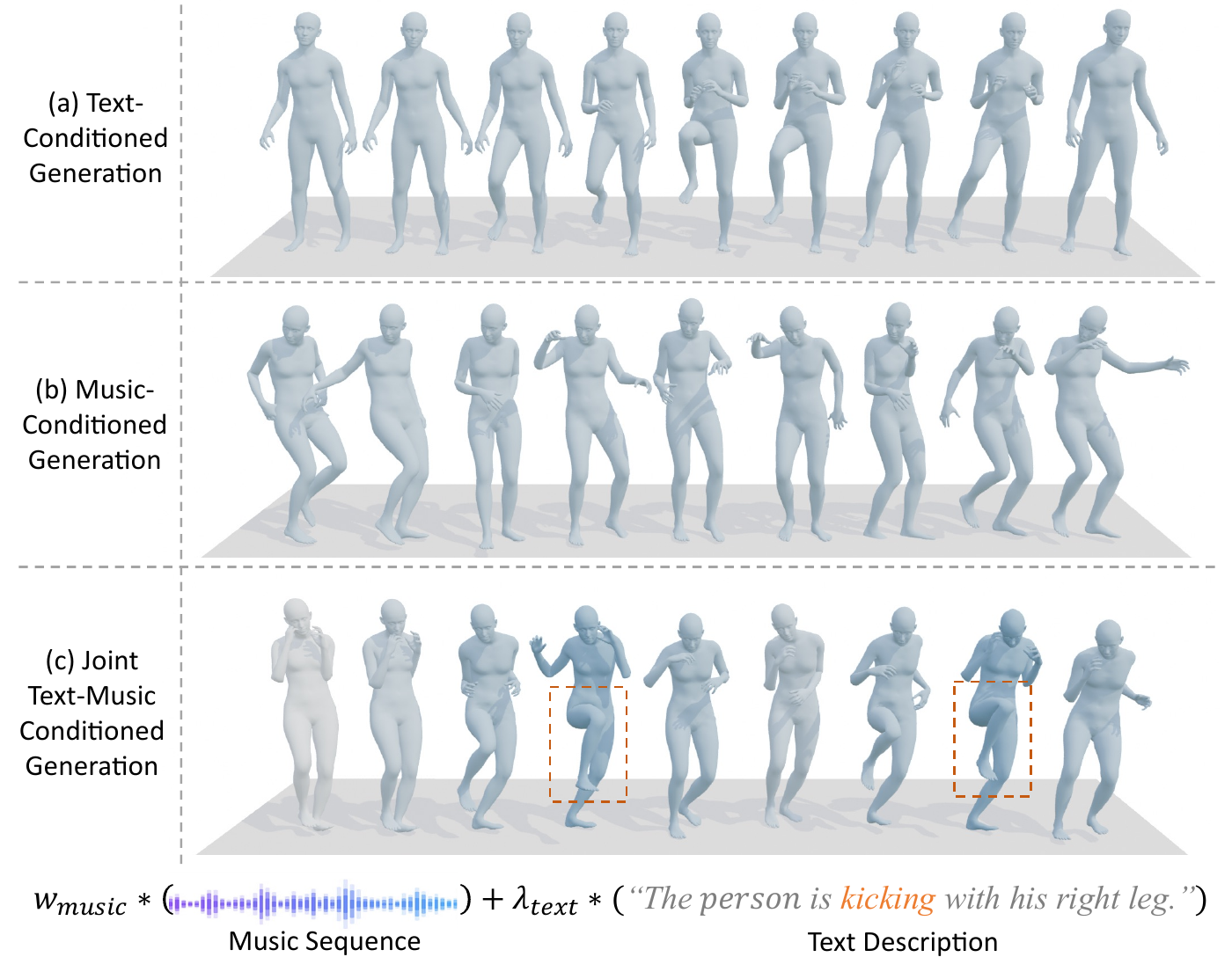}
  \caption{Text--music controllability at inference.}
  \label{fg:weight}
\vspace{-0.3cm}
\end{figure}

In inference, we apply the classifier-free guidance~\cite{ho2022classifier} to the music-conditioned backbone to continuously regulate the influence of music on the generated motion. Adjusting the music guidance scale yields a smooth continuum of behaviours, ranging from text-only generation under a null music condition, to music-only generation without text residual injection, and to joint text--music generation when both conditioning pathways are active. This inference mechanism provides a controllable trade-off between rhythmic fidelity and semantic steering, as illustrated in Figure~\ref{fg:weight}.


\begin{table*}[!t]
\caption{A quantitative comparison on FineDance. The best results are in \textbf{bold} and the second-best are \underline{underlined}. $\downarrow$ indicates lower is better, $\uparrow$ indicates higher is better, and  $\rightarrow$ indicates closer to the ground truth is better. $*$ marks abnormally high diversity values caused by discontinuous motions~\cite{li2021ai}.}
\label{table_1}
\centering
\footnotesize
\resizebox{0.99\textwidth}{!}{%
\begin{tabular}{cccccclc}
\toprule
& \multicolumn{2}{c}{\textbf{Motion Quality}} & \multicolumn{2}{c}{\textbf{Motion Diversity}} & \multirow{2}{*}{\textbf{PFC}$\downarrow$}  &\multirow{2}{*}{\textbf{PBC}$\rightarrow$  }& \multirow{2}{*}{\textbf{BAS}$\uparrow$} \\  
\cmidrule(lr{0.35em}){2-3} \cmidrule(lr{0.35em}){4-5}
& FID\_hand$\downarrow$ & FID\_body$\downarrow$ & Div\_hand$\uparrow$ & Div\_body$\uparrow$ &  && \\ 
 \midrule
 GT &  /& /& 11.82 ± 0.1314& 10.18 ± 0.1327 & /&5.23 ± 0.16& 0.2318 ± 0.0070  \\
\midrule
DanceRevolution~\cite{huang2020dance}      & 219.52 ± 18.32    & 99.83 ± 7.79      & 1.85 ± 0.60        & 4.49 ± 0.25         & 6.81 ± 0.81      &23.39 ± 2.03& 0.2104 ± 0.0057       \\
MNET~\cite{kim2022brand}                  & 195.56 ± 5.04     & 154.79 ± 2.80       & 6.79 ± 0.20     & 8.25 ± 0.39\textsuperscript{*}     & 2.98 ± 0.11        &12.21 ± 0.15 & 0.1792 ± 0.0014       \\
Bailando~\cite{siyao2022bailando}          & 55.60 ± 8.15      & 57.77 ± 6.01         & 6.40 ± 0.68     & 4.27 ± 0.43        & 0.34 ± 0.01        &3.09 ± 0.06 & 0.2152 ± 0.0028       \\
EDGE~\cite{tseng2023edge}                  & 25.37 ± 3.24     & 51.56 ± 3.62           & 8.29 ± 0.30   &  5.88 ± 0.32         & 0.21 ± 0.03        &7.78 ± 0.07&  0.2171 ± 0.0056\\
FineNet~\cite{li2023finedance}             & 26.88 ± 3.09&  \underline{23.59 ± 3.56}& 8.30 ± 0.45& 6.64 ± 0.28 & \textbf{0.12 ± 0.01} &3.35 ± 0.11& 0.2066 ± 0.0046\\
 DGFM~\cite{liu2024dgfm}& 20.699 ± 3.52& 24.63 ± 3.14& \underline{8.77 ± 0.41}& \underline{6.77 ± 0.75}& 0.20 ± 0.01& \underline{4.23 ± 0.06}&0.2153 ± 0.0054\\
 LODGE~\cite{li2024lodge} & \underline{18.36 ± 2.10}& 47.56 ± 1.37& 8.57 ± 0.36& 5.41 ± 0.27& \underline{0.13 ± 0.01}  & 3.46 ± 0.06&\underline{0.2327 ± 0.0050}\\
\midrule
 TeMuDance& \textbf{15.90 ± 3.28}& \textbf{23.41 ± 1.78}& \textbf{9.15 ± 0.37}& \textbf{6.89 ± 0.36}&  0.19 ± 0.01&\textbf{4.95 ± 0.10}& \textbf{0.2342 ± 0.0057}\\
\bottomrule
\end{tabular}}%
\end{table*}

\section{Experiments and Results}

\subsection{Experimental Setup}

\textbf{Implementation Details.}
We use motion and music sequences of $4$ seconds, corresponding to $N=120$ frames, and generate $52$-joint dance clips. The music-to-dance backbone is trained with Adan~\cite{xie2024adan} using a learning rate of $2\times10^{-4}$ and an $L_2$ reconstruction objective for $1000$ epochs with a batch size of $128$. For text-guided control fine-tuning, we train for $200$ epochs with a batch size of $96$. During inference, we use the standard DDPM sampler with $T=1000$ steps and classifier-free guidance with a scale of $3$.

\textbf{Datasets.}
Given that triplet-level supervision is not available for this task, we formulate the setting under disjoint supervision and leverage a music--dance dataset together with a text--motion dataset for training and evaluation.
For music--dance supervision, we use the FineDance~\cite{li2023finedance} dataset, which contains 14.6 hours of paired music and 52-joint 3D SMPL-X motions across 16 genres.
For text--motion supervision, we use the HumanML3D~\cite{guo2022generating} dataset, which contains 14,616 motions with 44,970 natural-language descriptions covering a diverse range of daily actions, each represented as SMPL-based 3D sequences.

\textbf{Evaluation metrics.}
We evaluate our method in terms of motion quality using the Fr\'echet Inception Distance (FID) between feature distributions of generated and real motions~\cite{li2021ai,li2020learning,heusel2017gans}, diversity using the diversity score adopted in Bailando~\cite{siyao2022bailando}, music--motion synchronisation using the Beat Alignment Score (BAS)~\cite{siyao2022bailando}, and physical plausibility using Physical Foot Contact (PFC) and Physical Body Contact (PBC)~\cite{tseng2023edge,luo2024popdg}.

To quantify text controllability under music conditioning, we propose Kinematic Primitive Success (KPS). Standard text--motion metrics such as R-Precision and matching score rely on a text--motion embedding space trained on HumanML3D and are designed for text-only generation. Applying them to our setting would require removing the music condition, fundamentally changing the task distribution and making the resulting scores unreliable for measuring text controllability under music conditioning. KPS directly evaluates whether textual conditioning drives the expected kinematic patterns in generated dances.
For each text prompt, we generate $R$ dance sequences conditioned on both the prompt and a randomly sampled music clip, and compare them against $R$ matched sequences generated with identical music and random seeds but empty text input. Each sequence is evaluated by a deterministic kinematic predicate defined for the target prompt, based on statistics such as relative joint heights, ground-plane displacement, and cumulative body rotation. We report the prompted success rate, the null-text success rate, and their difference as \textit{lift}. A positive lift indicates that textual conditioning actively induces the target motion pattern beyond what music alone produces.
To reduce variance from music selection, we repeat this protocol over $G$ independent groups, each with a different randomly sampled music clip, and report mean results. Results are aggregated into four families: pose-level, trajectory-level, rotation-level and temporal-level. The complete protocol and predicate definitions are provided in Appendix~\ref{app:kcs_predicates}.

\textbf{Baselines.} 
We consider two evaluation settings.
(i) Music-driven dance generation. We benchmark our method against representative and recent state-of-the-art music-conditioned dance generators on FineDance, including DanceRevolution~\cite{huang2020dance}, MNET~\cite{kim2022brand}, Bailando~\cite{siyao2022bailando}, EDGE~\cite{tseng2023edge}, FineNet~\cite{li2023finedance}, DGFM~\cite{liu2024dgfm}, and LODGE~\cite{li2024lodge}, following their standard evaluation protocols whenever available.
(ii) Text--music controlled generation. We qualitatively compare with TM2D~\cite{gong2023tm2d}, as it similarly combines music--dance and text--motion datasets to enable text and music conditioned dance generation.

\begin{figure*}[t]
  \centering
  \includegraphics[width=1\textwidth]{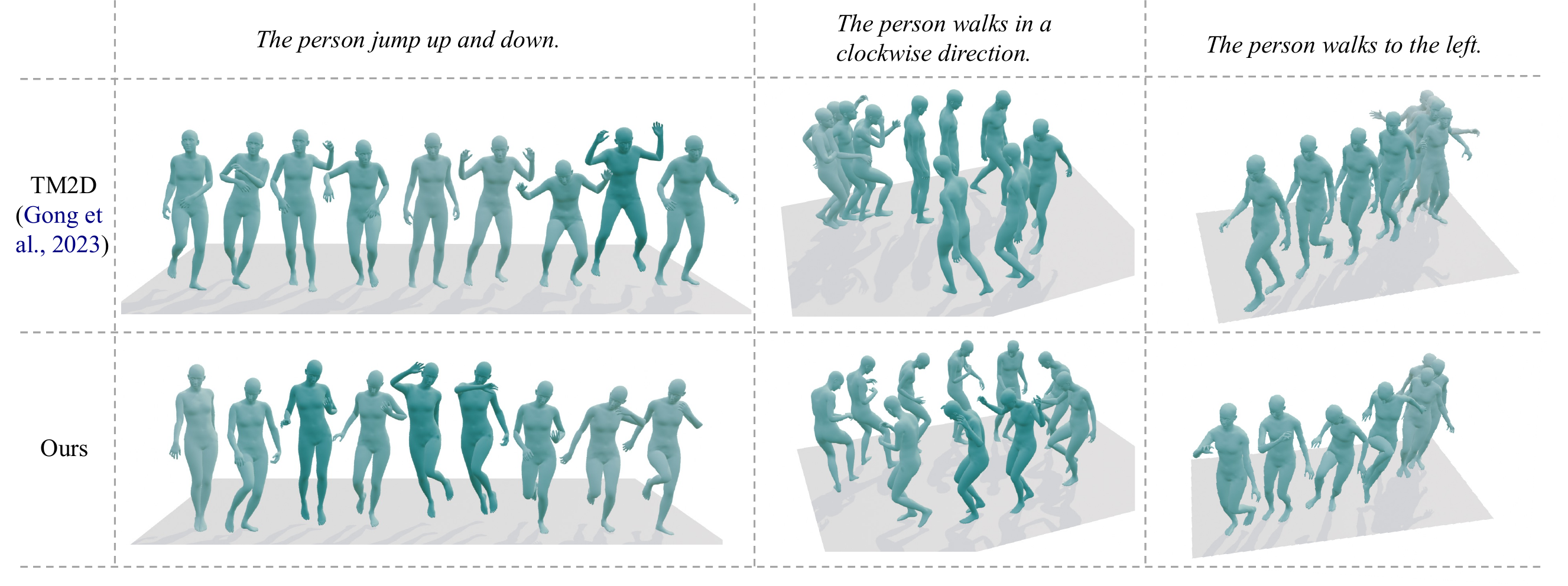}
  \caption{Visual comparison of the generated dance between the proposed method and TM2D~\cite{gong2023tm2d}. }
  \label{control}
\end{figure*}

\subsection{Results and Analysis}
\textbf{Evaluation on Music-Driven Dance Generation.} 
We validate music-driven dance generation on the FineDance test set, with results summarised in Table~\ref{table_1}. TeMuDance achieves the best overall motion quality, attaining the lowest FID for both hands and the body, indicating the closest match to the real-motion distribution. It also provides the strongest diversity on both hand and body motions. Beyond motion quality metrics, TeMuDance achieves the best physical body-contact score and the highest beat-alignment score, demonstrating that gains in realism and diversity are accompanied by improved physical plausibility and music–motion synchronisation. Although TeMuDance is not the top-performing method on PFC, it remains competitive and exhibits low foot-contact violations. Overall, TeMuDance demonstrates an excellent trade-off across realism, diversity, physical plausibility, and music--motion consistency.

\begin{figure}[!h]
  \centering
  \includegraphics[width=0.5\textwidth]{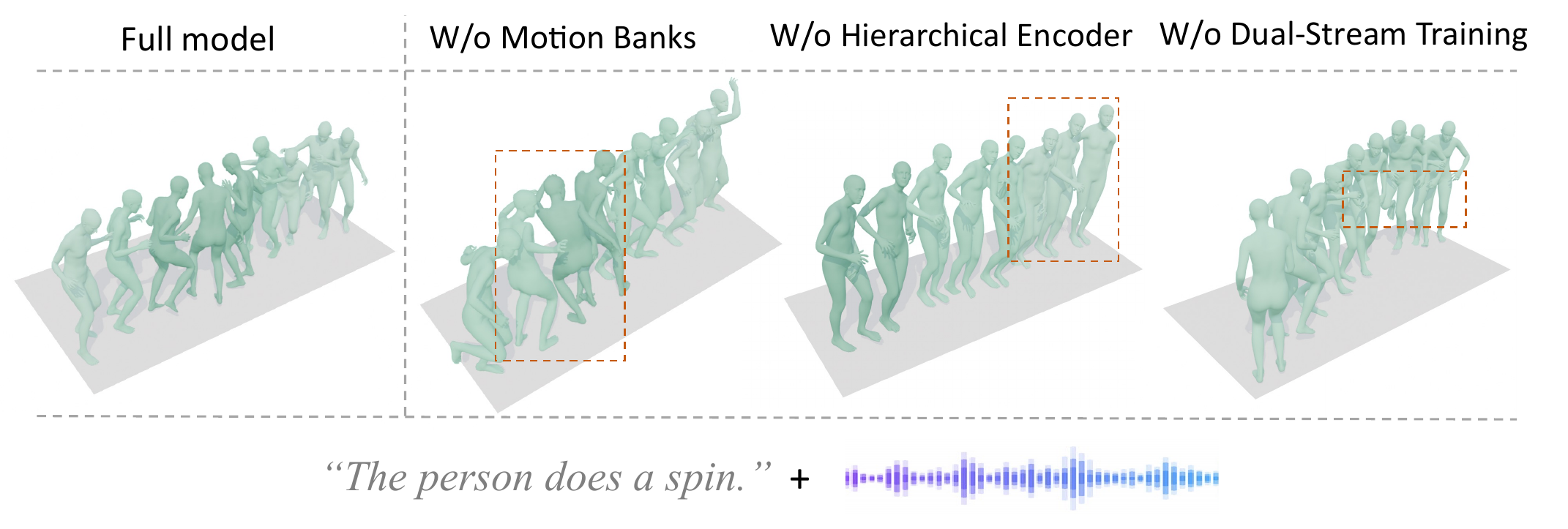}
  \caption{Visual comparisons of the ablation designs and our model.}
  \label{fg：abla}
  \vspace{-0.0cm}
\end{figure}

To further validate generalisation, we evaluate music-driven dance generation on AIST++~\cite{li2021ai}, a widely adopted benchmark in this domain. Table~\ref{tab:aistpp} summarises the results following the official evaluation protocol. TeMuDance achieves the second-best FID on both kinetic and manual features, the highest motion diversity, and competitive beat alignment, demonstrating that our backbone generalises well across different dance datasets.

\begin{table}[t]
    \centering
    \caption{Music-driven dance generation results on AIST++. Best in \textbf{bold}, second best \underline{underlined}.}
    \label{tab:aistpp}
    \resizebox{\linewidth}{!}{
        \begin{tabular}{lccccc}
            \toprule
            Method & FID$_k$ $\downarrow$ & FID$_m$ $\downarrow$ & Div$_k$ $\uparrow$ & Div$_m$ $\uparrow$ & BAS $\uparrow$ \\
            \midrule
            GT        & 17.10 & 10.60 & 8.19 & 7.45 & 0.2374 \\
            \midrule
            DanceNet~\cite{zhuang2022music2dance}  & 69.18 & 25.49 & 2.86 & 2.85 & 0.1430 \\
            Bailando~\cite{siyao2022bailando}  & \underline{28.16} & \textbf{9.62}  & \textbf{7.83} & \textbf{6.34} & 0.2332 \\
            DiffDance~\cite{qi2023diffdance} & \textbf{24.09} & 20.68 & 6.02 & 2.89 & \underline{0.2418} \\
            EDGE~\cite{tseng2023edge}      & 42.16 & 22.12 & 3.96 & 4.61 & 0.2334 \\
            LODGE~\cite{li2024lodge}       & 37.09 & 18.79 & 5.58 & 4.85 & 0.2423 \\
            \midrule
            TeMuDance                      & 32.14 & \underline{17.62} & \underline{5.92} & \underline{6.23} & \textbf{0.2427} \\
            \bottomrule
        \end{tabular}
    }
\vspace{-0.1cm}
\end{table}

\textbf{Evaluation on Text Controllability.}
Table~\ref{tab:kcs_primitive} reports per-primitive KPS results across eight text prompts, and Table~\ref{tab:kcs_family} summarises the family-level aggregates.
Across all families, prompted generations consistently outperform their null-text counterparts, confirming that textual conditioning actively steers the generated motion.
Trajectory-level control exhibits the largest family lift of $+60.0\%$, driven primarily by walk\_move at $+80.0\%$ and jump at $+40.0\%$, indicating that text provides strong locomotion and directional guidance.
Pose-level prompts achieve a family lift of $+42.5\%$, with kick and hands\_up reaching $+60.0\%$ and $+50.0\%$ respectively. These actions are almost entirely absent under null-text conditioning, suggesting that textual guidance can reliably introduce fine-grained postures that music alone does not produce.
Rotation-level and temporal families both achieve a lift of $+20.0\%$. Their relatively higher null rates indicate that turn and wave patterns already emerge to some extent from music conditioning, yet text still provides meaningful amplification.

\begin{table}[t]
    \centering
    \caption{Primitive-level KPS results. Prompt\%: success rate with text conditioning. Null\%: success rate with empty text. Lift\%: the difference indicating the effect of textual control.}
    \label{tab:kcs_primitive}
    \resizebox{\linewidth}{!}{
        \begin{tabular}{llccc}
            \toprule
            Family & Primitive & Prompt\% $\uparrow$ & Null\% & Lift\% $\uparrow$ \\
            \midrule
            \multirow{4}{*}{Pose-level}
            & crouch    & 90.0  & 50.0 & +40.0 \\
            & hands\_up & 50.0  &  0.0 & +50.0 \\
            & kick      & 60.0  &  0.0 & +60.0 \\
            & clap      & 20.0  &  0.0 & +20.0 \\
            \cmidrule(lr){1-5}
            \multirow{2}{*}{Trajectory-level}
            & walk\_move & 100.0 & 20.0 & +80.0 \\
            & jump       &  50.0 & 10.0 & +40.0 \\
            \cmidrule(lr){1-5}
            Rotation-level & turn  & 60.0 & 40.0 & +20.0 \\
            \cmidrule(lr){1-5}
            Temporal-level     & wave  & 60.0 & 40.0 & +20.0 \\
            \midrule
            \multicolumn{2}{l}{\textit{Macro-average}} & 61.3 & 20.0 & +41.3 \\
            \bottomrule
        \end{tabular}
    }
\end{table}

\begin{table}[t]
    \centering
    \caption{Family-level KPS results averaged from Table~\ref{tab:kcs_primitive}.}
    \label{tab:kcs_family}
    \resizebox{0.34\textwidth}{!}{
        \begin{tabular}{@{}l rrr@{}}
            \toprule
            Family & Prompt\% $\uparrow$ & Null\% & Lift\% $\uparrow$ \\
            \midrule
            Pose       & 55.00 & 12.50 & +42.50 \\
            Trajectory & 75.00 & 15.00 & +60.00 \\
            Rotation   & 60.00 & 40.00 & +20.00 \\
            Temporal   & 60.00 & 40.00 & +20.00 \\
            \bottomrule
        \end{tabular}

    }
\vspace{-0.3cm}
\end{table}

Table~\ref{tab:tradeoff} examines the interplay between text control strength and dance quality under varying guidance scales. The text column indicates whether textual conditioning is active, and the music column denotes the classifier-free guidance scale applied to the music-conditioned backbone. When music guidance is weaker, textual conditioning exerts a stronger influence on the generated motion, yielding higher KPS lift in the pose, trajectory, and rotation families. Temporal lift remains stable across scales, suggesting that periodic motion patterns such as waving are less sensitive to music guidance strength. Conversely, stronger music guidance improves beat alignment, as reflected by higher BAS, while still providing measurable controllability gains across most families. This confirms that users can smoothly navigate the trade-off between rhythmic fidelity and semantic steering by adjusting the guidance scale. We do not report FID in this analysis because action-level text control can deliberately shift the generated motion away from paired references, leading to uninformative FID values.
 
\begin{table}[t]
    \centering
    \caption{Controllability--quality trade-off under different guidance scales.}
    \label{tab:tradeoff}
    \resizebox{\linewidth}{!}{
        \begin{tabular}{cccccccccc}
            \toprule
            \multirow{2}{*}{Text} & \multirow{2}{*}{Music} & \multicolumn{4}{c}{Dance Quality} & \multicolumn{4}{c}{KPS Lift\% $\uparrow$} \\
            \cmidrule(lr){3-6} \cmidrule(lr){7-10}
            & & Div$_h$ $\uparrow$ & Div$_b$ $\uparrow$ & PFC $\downarrow$ & BAS $\uparrow$ & Pose & Traj. & Rot. & Temp.  \\
            \midrule
            0 & 1 & 7.93 & 5.12 & \textbf{0.12} & 0.2145 & --  & --  & -- & -- \\
            0 & 2 & 8.93 & 6.18 & 0.16 & 0.2160 & --  & --  & -- & -- \\
            0 & 3 & \textbf{9.15} & \textbf{6.89} & 0.19 & \textbf{0.2342} & --  & --  & -- & -- \\
            \midrule
            1 & 1 & 7.09 & 7.42 & 0.17 & 0.2041 & 47 & 65 & 30 & 20 \\
            1 & 2 & 5.31 & 6.18 & 0.19 & 0.2059 & 42 & 60 & 20 & 20 \\
            1 & 3 & 7.47 & 8.32 & 0.29 & 0.2141 & 35 & 30 & 10 & 20 \\
            \bottomrule
        \end{tabular}
    }
\vspace{-0.5cm}
\end{table}

\textbf{Qualitative Comparison on Music-text Conditioned  Generation.}
Figure~\ref{control} presents a qualitative comparison between TM2D~\cite{gong2023tm2d} and our method for joint music–text controlled dance generation. 
It shows that TM2D often separates the generated dance motion from the text-controlled action, making the instruction appear as an isolated segment rather than being fused into the choreography. For example, in the “clockwise direction” case, TM2D first generates several dance-like poses before briefly switching to a walking-and-turning pattern mid-sequence. This behaviour is consistent with the VQ-VAE-based discrete codebook representation in TM2D. In particular, quantised motion tokens from two datasets with different distributions are jointly used for training, which can encourage piecewise composition rather than continuous cross-modal fusion. In contrast, our method maintains the dance characteristics while enforcing the textual instruction throughout, resulting in more coherent joint control under combined music and text conditioning.

\subsection{Ablation Study}
To assess the necessity of each proposed component, we qualitatively compare the full model with three ablated variants.
As shown in Figure~\ref{fg：abla}, removing the retrieval mechanism significantly degrades generation quality. The resulting motion is lethargic and lacks rhythmic dynamism; instead of executing the requested "spin," the model produces a slow, partial rotation. This indicates that motion banks provide essential priors for both semantic controllability and beat-aligned dynamics.
Similarly, without the Hierarchical Encoder, the model struggles with precise semantic control. While the character attempts a turning motion, the execution is stiff and mechanically flawed, as highlighted by the orange box, lacking the fluidity and definition of a well-controlled action. This confirms that hierarchical modelling is necessary to enable fine-grained control over complex motion units. Finally, the model trained without the dual-stream strategy tends to over-prioritize textual instructions at the expense of dance fidelity. This suggests that the dual-stream strategy is vital for balancing strong semantic guidance with the inherent physical plausibility and coherence of music-driven dance.

Table~\ref{tab:abla_kcs} quantifies the contribution of each component using KPS lift. Removing the motion bank causes the most significant degradation across all four families, confirming that cross-modal retrieval provides essential priors for semantic steering. Removing the Hierarchical Encoder also reduces controllability, particularly at the rotation level.
 
\begin{table}[t]
    \centering
    \caption{Ablation on text controllability using KPS lift (\%).}
    \label{tab:abla_kcs}
    \resizebox{\linewidth}{!}{
        \begin{tabular}{lcccc}
            \toprule
            Setting & Pose $\uparrow$ & Traj. $\uparrow$ & Rot. $\uparrow$ & Temp. $\uparrow$ \\
            \midrule
            Full model               & \textbf{42} & \textbf{60} & \textbf{20} & \textbf{20}\\
            w/o Motion Bank           & 25 & 30 & 20 & 0 \\
            w/o Hierarchical Encoder  & 40 & 35 & 25& 20 \\
            w/o Dual-Stream Strategy    & 35 & 20 & 10 & 20\\
            \bottomrule
        \end{tabular}
    }
\end{table}

\subsection{User Study}

We conduct a user preference study with $20$ participants. For the general music-to-dance assessment, we randomly sample $12$ music clips from the test set. For the text-driven controllability task, we select $8$ test cases focusing on specific action instructions.
For dance generation quality, we obtain preference rates of $62.5$\%--$85.4$\%, indicating that participants generally favour the motions synthesized by our dance generation backbone in terms of fidelity and physical plausibility. Compared with TM2D, our model achieves a $75.0$\% preference on Choreographic Coherence, indicating more temporally continuous choreography with fewer clip-level composition discontinuities. Meanwhile, we maintain strong semantic controllability, suggesting that improved coherence does not come at the expense of instruction following.

\vspace{0.4cm}

\begin{table}[t]
    \centering
    \caption{Perceptual evaluation of generated samples}
    \label{tab:user_study}
    \resizebox{\linewidth}{!}{
        \begin{tabular}{lc}
            \toprule
            Comparison & Ours Win (\%) \\
            \midrule
            Music-to-Dance Generation Quality & \\
            \cmidrule(lr){1-2}
            \hspace{3mm} vs. Bailando~\cite{siyao2022bailando} & 85.4\\
            \hspace{3mm} vs. EDGE~\cite{tseng2023edge}         & 77.9\\
            \hspace{3mm} vs. FineDance~\cite{li2023finedance}   & 62.5\\
             \hspace{3mm} vs. LODGE~\cite{li2024lodge}          & 69.2\\
            \midrule
            Text-Driven Controllability  (vs. TM2D~\cite{gong2023tm2d}) & \\
            \cmidrule(lr){1-2}
            \hspace{3mm} Choreographic Coherence& 75.0\\
            \hspace{3mm} Semantic Controllability& 61.9\\
            \bottomrule
        \end{tabular}
    }
    
\end{table}

\section{Conclusion}
We presented \textit{TeMuDance}, a framework that enables fine-grained textual control for music-driven 3D dance generation without requiring paired music--text--motion triplets. By bridging disjoint music--dance and text--motion datasets through motion-centred contrastive alignment, TeMuDance transfers semantic concepts to a frozen music-conditioned generator via a lightweight control branch. We further proposed Kinematic Primitive Success, a task-aligned metric that directly measures whether textual conditioning induces the expected kinematic patterns. Experiments demonstrated that TeMuDance achieves effective text controllability while maintaining competitive dance quality and rhythmic fidelity.


\bibliographystyle{ACM-Reference-Format}
\bibliography{software}

\clearpage

\appendix

\section{Detailed Motion Representation}
\label{app:motion}
In this section, we detail the composition of the motion representation $\bm{x} \in \mathbb{R}^{k \times F}$ ($F=319$) derived from the SMPL-X parameterisation~\cite{loper2023smpl}.
The feature vector comprises three parts: ($1$) Joint Rotations: The poses of $52$ skeletal joints are transformed into a continuous $6$-dimensional rotation representation, yielding a $312$-dimensional vector; ($2$) Root Translation: A $3$-dimensional vector representing the global trajectory in world space; and ($3$) Foot Contact: Following~\cite{tseng2023edge}, we append a $4$-dimensional binary signal encoding the heel and toe contact states. Together, these components constitute the final feature dimension of $312 + 3 + 4 = 319$.

\section{Retrieval Quality Analysis}
\label{app:retrieval}
Our retrieval operates as motion-to-motion nearest-neighbour search in the learned shared embedding space, rather than direct cross-modal text-to-music matching. For each sample, we retrieve the nearest motion from the other dataset and attach the retrieved sample's paired modality to form a pseudo triplet.

Table~\ref{app_sim} summarises the cosine similarity distribution of raw top-1 retrieval results before thresholding. With a threshold of $0.8$, the acceptance rates are $69.15\%$ for FineDance to HumanML3D and $89.51\%$ for HumanML3D to FineDance, as shown in Table~\ref{tab:retrieval_accept}. All rejected matches are replaced by a null condition, preventing low-confidence pseudo annotations from propagating into training.

\begin{table}[!htbp]
    \centering
    \caption{Cosine similarity distribution of top-1 retrieval results before thresholding.}
    \label{app_sim}
    \resizebox{\linewidth}{!}{
        \begin{tabular}{lcccccc}
            \toprule
            Direction & Min & P10 & Median & P90 & Max & Mean \\
            \midrule
            FineDance $\rightarrow$ HumanML3D& 0.601 & 0.765 & 0.820 & 0.863 & 0.933 & 0.816 \\
            HumanML3D $\rightarrow$ FineDance & 0.727 & 0.829 & 0.872 & 0.902 & 0.933 & 0.868
 \\
            \bottomrule
        \end{tabular}
    }
\end{table}

\begin{table}[!htbp]
    \centering
    \caption{Retrieval acceptance rates with cosine similarity threshold $0.8$.}
    \label{tab:retrieval_accept}
    \resizebox{\linewidth}{!}{
        \begin{tabular}{lcc}
            \toprule
            Direction & Acceptance Rate (\%) & Null Replaced (\%) \\
            \midrule
            FineDance $\rightarrow$ HumanML3D & 69.15 & 30.85 \\
            HumanML3D $\rightarrow$ FineDance & 89.51 & 10.49 \\
            \bottomrule
        \end{tabular}
    }
\end{table}

\section{Kinematic Predicate Definitions}
\label{app:kcs_predicates}
 
All predicates operate on 22-joint 3D positions at 30 FPS. The height axis is automatically identified as the axis along which the head-to-pelvis offset is maximally positive, and the ground-plane axes are the remaining two dimensions. Table~\ref{tab:kcs_predicates} lists each predicate with its family assignment, kinematic criterion, and threshold.

The hip orientation vector is defined as the ground-plane projection from the left hip to the right hip joint. Yaw is computed as the arctangent of this vector, and cumulative yaw is the sum of absolute frame-to-frame angular differences with angles wrapped to $[-\pi, \pi]$.
For walk\_move, step crossings are counted as the number of sign changes in the along-movement projection of the left-right ankle displacement, capturing the alternation characteristic of a walking gait.
For jump, the upward velocity gate requires the peak frame-to-frame pelvis velocity along the height axis to exceed a threshold, suppressing false positives from slow vertical drift.
For kick, the dominance gap is the difference between the larger and smaller ankle lifts, ensuring that only single-leg raises are classified as kicks rather than symmetric bounces.
For the wave predicate, the dominant frequency is obtained via FFT of the Hanning-windowed wrist-shoulder relative displacement along the axis of greatest range.

\begin{table*}[t]
    \centering
    \caption{Kinematic predicate definitions for KPS evaluation. All spatial thresholds are in the coordinate units of the SMPL skeleton. $h$: height axis. $\sigma_s$: mean ground-plane shoulder width.}
    \label{tab:kcs_predicates}
    \resizebox{\textwidth}{!}{
        \begin{tabular}{llll}
            \toprule
            Primitive & Family & Kinematic Criterion & Threshold \\
            \midrule
            walk\_move & Trajectory & Ground-plane pelvis displacement and left-right foot alternation & disp $> \max(0.25,\; 1.0 \times \sigma_s)$ and $\geq 2$ step crossings \\
            jump & Trajectory & Peak pelvis height minus mean pelvis height over first 10 frames, with upward velocity gate & lift $> 0.12$ and peak velocity $> 0.6$ \\
            turn & Rotation & Cumulative absolute yaw change of the hip orientation vector & $> 90^{\circ}$ \\
            crouch & Pose & Mean pelvis height over first 10 frames minus minimum pelvis height & $> \max(0.08,\; 0.15 \times |h_{\text{rest}}|)$ \\
            hands\_up & Pose & Fraction of frames where wrist height exceeds shoulder height & $\geq 8\%$ for both, or $\geq 16\%$ for either \\
            kick & Pose & Maximum ankle lift above mean ankle height over first 10 frames, with single-leg dominance & lift $> 0.30$ and dominance gap $> 0.08$ \\
            clap & Pose & Fraction of frames where wrist-to-wrist distance falls below $0.60 \times \sigma_s$ & $\geq 10\%$ \\
            wave & Temporal & Wrist-shoulder relative motion amplitude, direction changes, and dominant frequency & amp $> 0.40 \times \sigma_s$, $\geq 3$ zero-crossings, freq $\in [0.5, 2.5]$ Hz \\
            \bottomrule
        \end{tabular}
    }
\end{table*}

\end{document}